\documentclass[runningheads]{llncs}
\usepackage[T1]{fontenc}
\usepackage{graphicx}
\usepackage{booktabs}
\usepackage[misc]{ifsym}

\usepackage{mwe}
\usepackage{amsmath, amssymb}
\usepackage{multirow}
\usepackage{diagbox}
\usepackage{pifont}
\usepackage{subfigure}

\begin{document}
\tocauthor{Nan Sun, Xixun Lin, Zhiheng Zhou, Yanmin Shang, Zhenlin Cheng, Yanan Cao}
\toctitle{Evidential Spectrum-Aware Contrastive Learning for OOD Detection in Dynamic Graphs} 

\title{Evidential Spectrum-Aware Contrastive Learning for OOD Detection in Dynamic Graphs}

\titlerunning{EviSEC for OOD Detection in Dynamic Graphs}

\author{Nan Sun\inst{1,2} \and
Xixun Lin\inst{1,2} \Letter \and
Zhiheng Zhou\inst{3} \and\\
Yanmin Shang\inst{1,2} \and
Zhenlin Cheng\inst{4} \Letter \and
Yanan Cao\inst{1,2}}


\authorrunning{Nan Sun et al.}
\institute{Institute of Information Engineering, Chinese Academy of Sciences, Beijing, China \email{\{sunnan, linxixun, shangyanmin, caoyanan\}@iie.ac.cn}
\and
School of Cyber Security, University of Chinese Academy of Sciences, Beijing, China 
\and
Academy of Mathematics and Systems Science, Chinese
Academy of Sciences \\\email{zhouzhiheng@amss.ac.cn}
\and
Beijing Wuzi University, Beijing, China \\\email{chengzhenlin@bwu.edu.cn}
}
\maketitle              

\begin{abstract}
Recently, Out-of-distribution (OOD) detection in
dynamic graphs, which aims to identify whether incoming data deviates from the distribution of the in-distribution (ID) training set, has garnered considerable attention in security-sensitive fields. Current OOD detection paradigms primarily focus on static graphs and confront two critical challenges: i) high bias and high variance caused by single-point estimation, which makes the predictions sensitive to randomness in the data; ii) score homogenization resulting from the lack of OOD training data, where the model only learns ID-specific patterns, resulting in overall low OOD scores and a narrow score gap between ID and OOD data. To tackle these issues, we first investigate OOD detection in dynamic graphs through the lens of Evidential Deep Learning (EDL). Specifically, we
propose \textbf{EviSEC}, an innovative and effective OOD detector via \underline{\textbf{Evi}}dential \underline{\textbf{S}}pectrum-awar\underline{\textbf{E}} \underline{\textbf{C}}ontrastive Learning. 
We design an evidential neural
network to redefine the output as the posterior Dirichlet distribution, explaining the randomness of inputs through the uncertainty of distribution, which is overlooked by single-point estimation. Moreover, spectrum-aware augmentation module generates OOD approximations to identify patterns with high OOD scores, thereby widening the score gap between ID and OOD data and mitigating score homogenization.
Extensive experiments on real-world datasets demonstrate that EviSAC effectively detects OOD samples in dynamic graphs. Our source code is available at \url{https://github.com/Sunnan191/EviSEC}.

\keywords{Dynamic graph \and Out-of-distribution detection \and Evidential deep learning \and Graph spectrum.}
\end{abstract}

\begin{figure}[htbp]
    \centering
    \includegraphics[width=10.5cm]{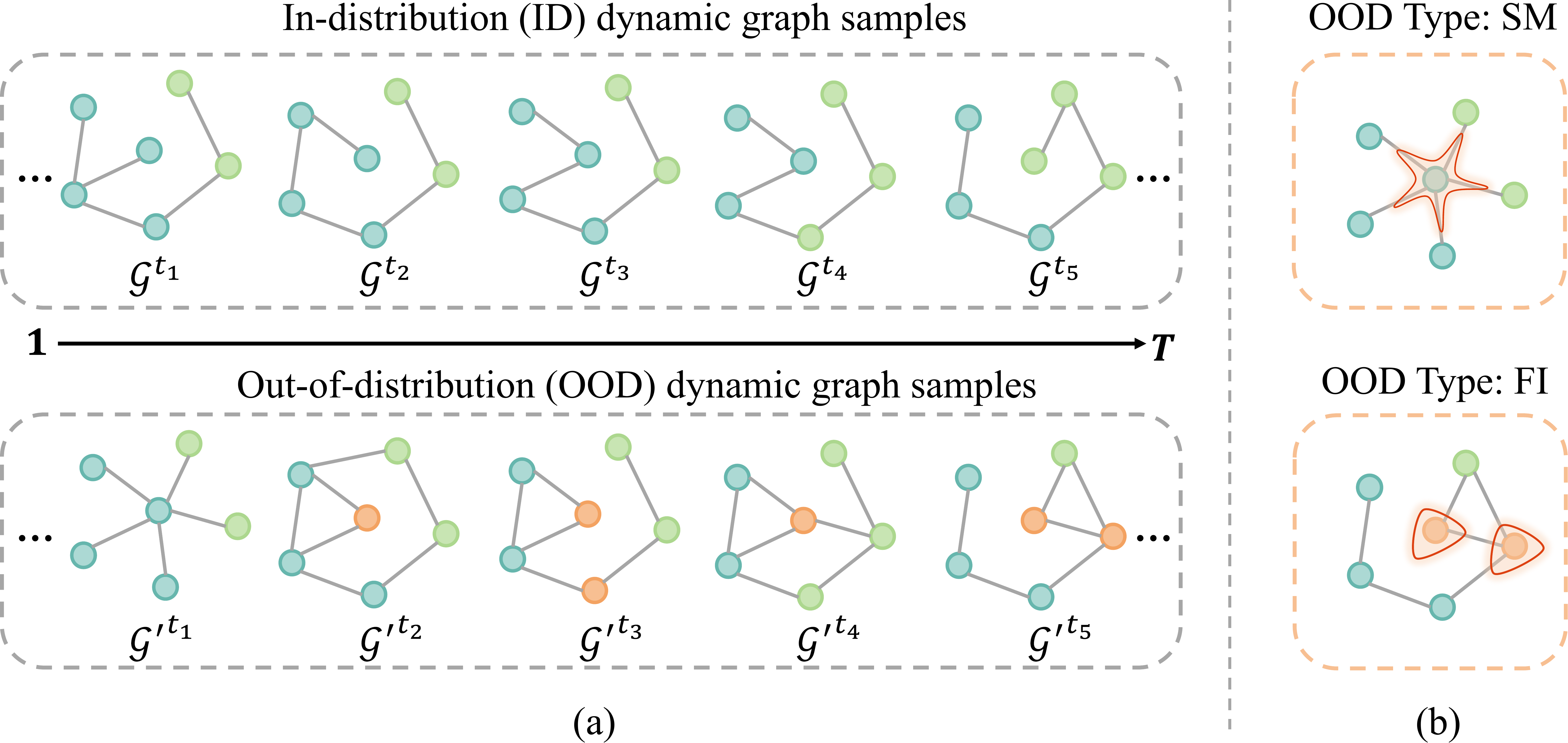} 
    \caption{Dynamic graph examples of (a) ID samples maintain consistent structures, while OOD samples exhibit deviations. (b) Examples include cases such as SM (Structure Manipulation) and FI (Feature Interpolation), highlighting structural changes and feature variations, respectively.}
    \label{fig:toy}
\end{figure}
\section{Introduction}

Real-world graph data often evolves temporally, allowing dynamic graphs to be ubiquitously applied across non-Euclidean domains such as citation networks~\cite{tang2008arnetminer}, social communities~\cite{nr}, and transaction records~\cite{kumar2018rev2}. Recently,
studies on dynamic graphs have gained increasing popularity, among which Dynamic Graph Neural Networks (DGNNs)~\cite{zhang2024temporal} have rapidly emerged as powerful approaches for dynamic graph representation learning~\cite{khoshraftar2024survey}.
These models primarily leverage Graph Neural Networks (GNNs) and sequence-based architectures to capture temporal variations in dynamic graphs.
\par

Most existing models on dynamic graphs are predominantly trained with the closed-world assumption that the training and test data share the same data distribution, with such data termed In-Distribution (ID) data. 
Nevertheless, real-world scenarios frequently deviate from this ideal assumption, often involving Out-of-Distribution (OOD) dynamic graphs, which are unobserved during the training process. Fig.~\ref{fig:toy} provides examples illustrating the behavior of ID and OOD dynamic graph with two OOD types. During the inference stage, performing predictions blindly without recognizing OOD samples can render the output unreliable and pose serious risks. Such cases are especially unacceptable in safety-critical domains, such as medical diagnostics~\cite{kononenko2001machine} and autonomous driving~\cite{yurtsever2020survey}. An ideal model should effectively handle both ID test performance and OOD detection performance. Therefore, OOD detection in dynamic graphs has seen a growing demand, which aims to identify whether incoming data deviate from the distribution of the training set.
\par

Early works in OOD detection primarily focus on static graphs. Liu et al.~\cite{liu2023good} propose a hierarchical contrastive learning method that captures common patterns of ID graphs across different granularities (node, graph, and group levels), so that OOD graphs that deviate from these patterns can be effectively identified. GNNSAFE~\cite{wu2023energy} extends energy-based models~\cite{ranzato2007unified} to static graphs and develops an energy function that classifies nodes with high energies as OOD samples. In parallel, research on anomaly detection~\cite{song2025uniform} in dynamic graphs has also gained attraction. Netwalk~\cite{yu2018netwalk} uses random walks and autoencoders to generate similar node representations for ID data, thereby identifying anomalous interactions between nodes with distinct representations. TADDY~\cite{liu2021anomaly} employs transformers to capture both global and local structural anomalies in node representations.
\textit{Although the above approaches perform fairly well, they are confronted with two critical limitations:}

i) \textbf{High bias and high variance caused by single-point estimation.} Most of the aforementioned methods follow a manner of single-point estimation, which overlook the inherent randomness presented in the data, so that their OOD detection results are sensitive to natural noise, resulting in high bias and high variance in the output.  

ii) \textbf{Score homogenization resulting from the lack of OOD training data.} During the training stage, only ID data is available whereas OOD data remains unseen. This will cause the problem of score homogenization, i.e., the model only learns the ID-specific patterns, making it tend to assign overall lower OOD scores, regardless of whether the sample is ID or OOD. The small score gap between ID and OOD data makes it challenging to effectively distinguish OOD samples.
\par
To address these limitations, we thoroughly investigate OOD detection in dynamic graphs from a novel perspective drawing on Evidential Deep Learning (EDL)---a theoretically grounded framework for uncertainty quantification through Dirichlet distributions. Specifically, we propose \textbf{EviSEC}, an innovative and effective OOD detector via \underline{\textbf{Evi}}dential \underline{\textbf{S}}pectrum-awar\underline{\textbf{E}} \underline{\textbf{C}}ontrastive Learning. We design an evidential neural network to reshape the output class probability of single-point estimation into a probability Dirichlet distribution, allowing us to describe the variability and randomness of the data through the uncertainty of distribution. Furthermore, we develop two loss functions to guide the model to output sharper Dirichlet distributions with low uncertainty scores for ID samples while preserve the ID performance. To tackle the issue of missing OOD samples, we propose a spectral-aware data augmentation module that generates OOD approximations of dynamic graphs. Based on this, the contrastive loss can enable the model to learn and assign higher uncertainty scores, thereby mitigating the problem of score homogenization. Finally, our OOD detector computes the learned uncertainty as the OOD score, which efficiently distinguishes OOD samples with higher uncertainty.
The main contributions of this study are summarized as follows: 
\begin{enumerate}
    \item EviSEC is the first to establish a direct link between EDL and OOD detection in dynamic graphs. We propose an evidential neural network that uses the uncertainty of posterior distribution to describe the randomness of the input, addressing the problem of single-point estimation. Our method effectively improves OOD detection while maintaining ID testing performance.
    \item We propose a spectral-aware data augmentation module to generate OOD approximations for dynamic graphs, alleviating the lack of OOD training samples. Based on this, our contrastive learning module enlarges the score gap, mitigating the issue of score homogenization and effectively enhancing the OOD detection performance.
    \item Experimental results on multiple real-world datasets validate the claimed advantages of our approach, which achieves consistent performance gains over multiple powerful competitors with an average AUROC improvement of 24.32\%.
\end{enumerate}
\section{Related Works}
\subsection{Dynamic Graph Representation Learning}
Early attempts in dynamic graph representation learning~\cite{khoshraftar2024survey,lin2021disentangled} employ traditional methods including random walks~\cite{grover2016node2vec}, matrix factorization~\cite{li2017attributed}, and temporal point processes~\cite{zuo2018embedding,zhang2024neural} to model graph information over time. However, compared to these methods, DGNNs outperform in expressive power, since they combine message passing and temporal modeling to better capture topological and temporal dynamics. For instance, EvolveGCN~\cite{pareja2020evolvegcn} employs RNNs to dynamically adapt the weights of GNNs across temporal steps. DEFT~\cite{DEFT2023} utilizes transformers to model temporal dependencies through self-attention mechanisms. Instead of using RNNs, LEDG~\cite{LEDG2022} applies gradient-based meta learning to learn updating strategies. While these methods excel in predictive tasks, they largely focus on ID data, often neglecting OOD detection in dynamic graphs. 

\subsection{Graph Out-of-distribution Detection}
Graph OOD detection primarily focuses on static graphs and can be broadly categorized into the following three approaches:
\paragraph{\textbf{Post-processing Methods.}} These model-agnostic methods directly process outputs of pretrained models to estimate OOD scores. For instance, Lee et al.~\cite{lee2018simple} use Mahalanobis distance to measure the deviation of input data. Similarly, maximum softmax probabilities (MSP)~\cite{hendrycks2016baseline} can be utilized as confidence scores, with lower values signaling OOD samples. ODIN~\cite{ODIN2017} enhances MSP by incorporating temperature scaling and input perturbation to increase the score gap. Yet, these methods typically follow single-point estimation, which is prone to high variance and bias, thereby hindering their ability to recognize the OOD patterns.
\paragraph{\textbf{Energy-based Methods.}}
These approaches are based on the framework of energy-based models, where energy scores measure the discrepancy between input data and the model's learned distribution. Specifically, energy scores~\cite{liu2020energy} effectively separate OOD samples from ID data, as OOD samples typically yield higher values. GNNSAFE~\cite{wu2023energy} develops an energy function directly extracted from GNNs trained with energy propagation on static graphs. Nevertheless, directly adapting these approaches to dynamic settings is challenging since they require a fundamental redesign of the energy propagation mechanism over time.
\paragraph{\textbf{Uncertainty-based Methods.}}
Uncertainty~\cite{abdar2021review} measures the unreliability of a model's predictive distribution. A previous study~\cite{lin2024graph} adopts the graph neural stochastic diffusion framework to model uncertainty in node classification tasks. Zhao et al.~\cite{zhao2020uncertainty} propose a graph-based kernel estimation method to predict node-level Dirichlet distributions within the EDL framework. DAEDL~\cite{yoon2024uncertainty} promotes EDL performance through a density estimation algorithm. In addition, several studies~\cite{zhang2025conformal,lin2025conformal} utilize the uncertainty measurement framework of conformal prediction to provide statistical guarantees for the detection results. However, the absence of OOD samples can cause severe score homogenization in detection.
\section{Preliminary}\label{preliminary} 
\subsection{Problem Definition}\label{sec:problem}
The dynamic graphs are interpreted as a sequence of graphs $\mathcal{G}^{1:T} = \{ \mathcal{G}^t \}_{t=1}^{T}$, where $T$ specifies the total number of timesteps. Each discrete graph snapshot $\mathcal{G}^t = (\mathcal{V}^t, \mathcal{E}^t, \mathbf{X}^t, \mathbf{A}^t)$ contains node set $\mathcal{V}^t$, edge set $\mathcal{E}^t$, adjacency matrix $\mathbf{A}^t \in \{0,1\}^{N^t \times N^t}$, and feature matrix $\mathbf{X}^t \in \mathbb{R}^{N^t \times d}$, where $N^t = |\mathcal{V}^t|$ and $d$ respectively denote the node count and the feature dimension at timestep $t$. 
Dynamic graph representation learning aims to develop a powerful encoder $f_\theta(\cdot,\cdot)$ to capture the temporal dependency. To achieve this, the graph snapshots $\mathcal{G}^{t:t + \Delta t}$ are commonly employed as input to model the dynamics within the temporal window $\Delta t$. Based on such input settings, we formulate OOD detection in dynamic graphs as the task of ascertaining whether $\mathcal{G}^{t:t + \Delta t}$ follow the same distribution as ID data:
\par
\begin{definition}[\textbf{OOD detection in Dynamic Graphs}]\label{definition_1_dynamic}
Let the ID dataset $\mathcal{D}_{\rm{in}}$ $($resp. OOD dataset $\mathcal{D}_{\text{out}}$$)$ consist of graph sequences drawn from the distributions $\mathbb{P}_{\text{in}}$ $($resp. $\mathbb{P}_{\text{out}}$$)$. The training dataset 
$\mathcal{D}^{\text{tr}} = \left\{ \mathcal{G}^t_{\text{in}} \right\}_{t=1}^{T}$ 
is a subset of $\mathcal{D}_{\text{in}}$, containing ID graphs at different timesteps. The test dataset $\mathcal{D}^{\text{te}}$ is constituted by two disjoint subsets, $\mathcal{D}^{\text{te}}_{\text{in}} \subset \mathcal{D}_{\text{in}}$ and $\mathcal{D}^{\text{te}}_{\text{out}} \subset \mathcal{D}_{\text{out}}$, i.e., $\mathcal{D}^{\text{te}} = \mathcal{D}^{\text{te}}_{\text{in}} \cup \mathcal{D}^{\text{te}}_{\text{out}}$, 
such that $\mathcal{D}^{\text{tr}} \cap \mathcal{D}_{\text{in}}^{\text{te}} = \emptyset$. For arbitrary input snapshots $\mathcal{G}^{t:t + \Delta t} \in \mathcal{D}^{\text{te}}$, the goal of OOD detection in dynamic graphs is to design a discriminant function $G(\cdot,\cdot)$ to determine whether $\mathcal{G}^{t:t + \Delta t}$ follow  $\mathbb{P}_{\text{in}}$ or $\mathbb{P}_{\text{out}}$ based on the OOD detection score:
\begin{equation}\label{detection}
\text{detect}(\mathcal{G}^{t:t + \Delta t}) = 
\begin{cases} 
1 & G\left(f_\theta(\mathcal{G}^{t:t + \Delta t}, \mathcal{W}), \mathcal{D}^{\text{tr}}\right) \geq \gamma, \\
0 & G\left(f_\theta(\mathcal{G}^{t:t + \Delta t}, \mathcal{W}), \mathcal{D}^{\text{tr}}\right) < \gamma,
\end{cases}
\end{equation}
where $ \gamma $ is the detection threshold, $\mathcal{W}$ represents the parameters of the encoder.
\end{definition}

\subsection{Evidential Deep Learning}\label{sec:edl}
\begin{figure}[h]
    \centering
    \includegraphics[width=8.5cm]{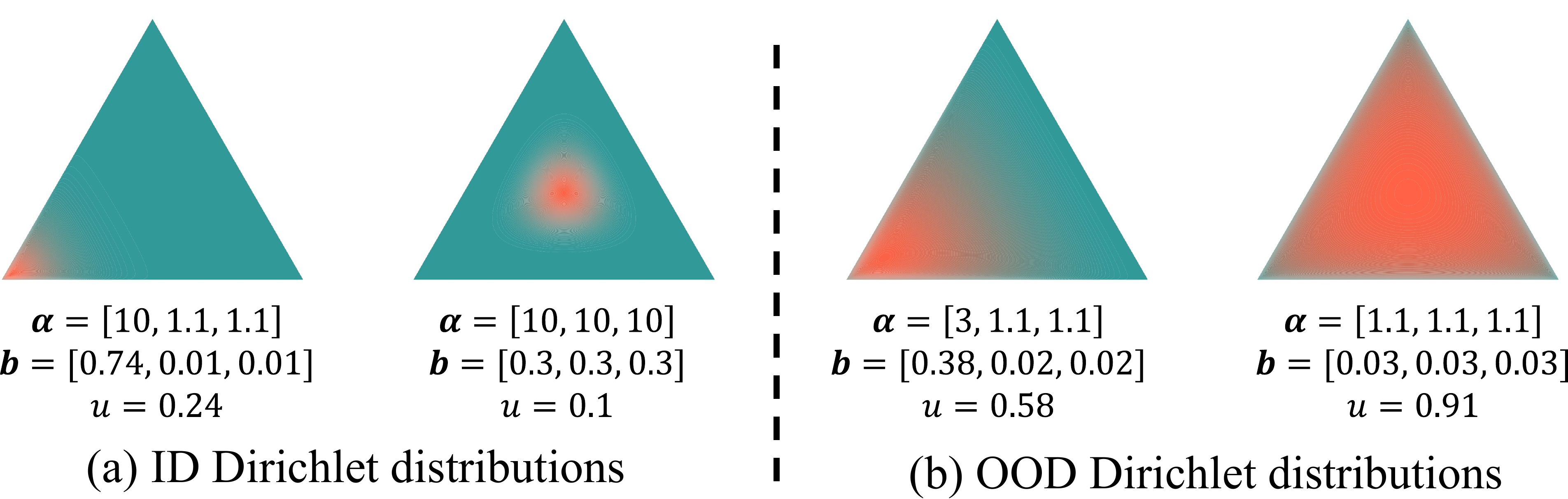} 
    \caption{Heatmaps of Dirichlet distributions in 3-class classification with four groups of concentration parameters and corresponding subjective opinions. Warm (resp. cool) colors represent relatively high (resp. low) probability density values in the distribution.}
    \label{fig:diri}
\end{figure}

EDL is grounded in Subjective Logic (SL) theory~\cite{josang2016subjective}, which represents subjective multinomial opinion as a non-negative triplet $\boldsymbol{\tau} = (\boldsymbol{b}, u, \boldsymbol{\beta})$. In a $K$-class classification problem, $\boldsymbol{b}=[b_1,b_2,\dots,b_K]$ assigns belief mass $b_i$ to each class; $u$ quantifies the overall uncertainty across classes, satisfying $u + \sum_{i=1}^K b_i = 1$; and  $\boldsymbol{\beta}=[\beta_1,\beta_2,\dots,\beta_K]$ is a predefined base rate vector. Based on $\boldsymbol{\tau}$, the projected probability distribution for each class is defined as $p_i = b_i + \beta_i u$, where $i = 1,2,\dots,K$. 
EDL adopts the K-dimensional Dirichlet distribution $\text{Dir}^{pr}(\boldsymbol{p}; w\boldsymbol{\beta})$ as the prior distribution, where $w$ is
the prior weight. The conjugate posterior of EDL, i.e., $\text{Dir}^{po}(\boldsymbol{p}; \boldsymbol{\alpha})$  with concentration (parameter) $\boldsymbol{\alpha} = [\alpha_1, \alpha_2, \cdots, \alpha_K]$ is defined as:
\begin{equation}\label{dir}
\text{Dir}^{po}(\boldsymbol{p}; \boldsymbol{\alpha}) = \frac{\Gamma\left(\sum_{i =1}^K \alpha_i\right)}{\prod_{i =1}^K \Gamma(\alpha_i)} \prod_{i =1}^K p_i^{\alpha_i - 1},
\end{equation}
where $\Gamma$ denotes the Gamma function, $\alpha_i \geq 0$, and $p_i \neq 0$ if $\alpha_i < 1$. To obtain the posterior distribution in Eq.~\ref{dir}, EDL first designs a post-processing module to capture the evidence vector $\boldsymbol{e} = [e_1, e_2, \cdots, e_K]$, which represents the observation evidences for $K$ classes. Then EDL combines the observed evidence with the prior to update the posterior concentration $\boldsymbol{\alpha}= [\alpha_1, \alpha_2, \cdots, \alpha_K]$ as
\begin{equation}
\alpha_i = e_i + \beta_i w, \text{where}\ i = 1,2,\dots,K. 
\end{equation}
Thereby the multinomial opinion  $\boldsymbol{\tau} = (\boldsymbol{b}, u, \boldsymbol{\beta})$ in SL can be equivalently represented by
EDL through the bijection mapping $F$ between $\boldsymbol{\tau}$ and $\text{Dir}^{po}(\boldsymbol{p}; \boldsymbol{\alpha})$:
\begin{equation}\label{F}
F: \boldsymbol{\tau} \longleftrightarrow \text{Dir}^{po}(\boldsymbol{p}; \boldsymbol{\alpha}), \text{where } b_i = \frac{\alpha_i-\beta_i w}{\alpha_\text{sum}},u=\frac{w\sum_{i=1}^K\beta_i }{\alpha_\text{sum}} ,\alpha_\text{sum} = \sum_{i=1}^K \alpha_i.
\end{equation}
EDL treats the predictive uncertainty $ u $ in Eq.~\ref{F} as a discriminative metric for OOD detection. Specifically, for ID samples shown in Fig.~\ref{fig:diri}(a), the model assigns relatively high evidence to at least one class, resulting in a sharp Dirichlet distribution (large $\alpha_\text{sum}$) with lower uncertainty ($ u\to0 $). Conversely, for OOD samples shown in Fig.~\ref{fig:diri}(b), the evidence allocated across all classes is relatively low, leading to a flat Dirichlet distribution (small $\alpha_\text{sum}$) with high uncertainty ($ u \to 1 $). This difference in uncertainty values clearly shows how the model can tell apart ID and OOD samples. A predefined threshold $ \gamma $ (e.g., $ \gamma = 0.5 $) can thus be applied to trigger OOD detection.


\section{METHODOLOGY}
\begin{figure}[t]
    \centering
    \includegraphics[width=\textwidth]{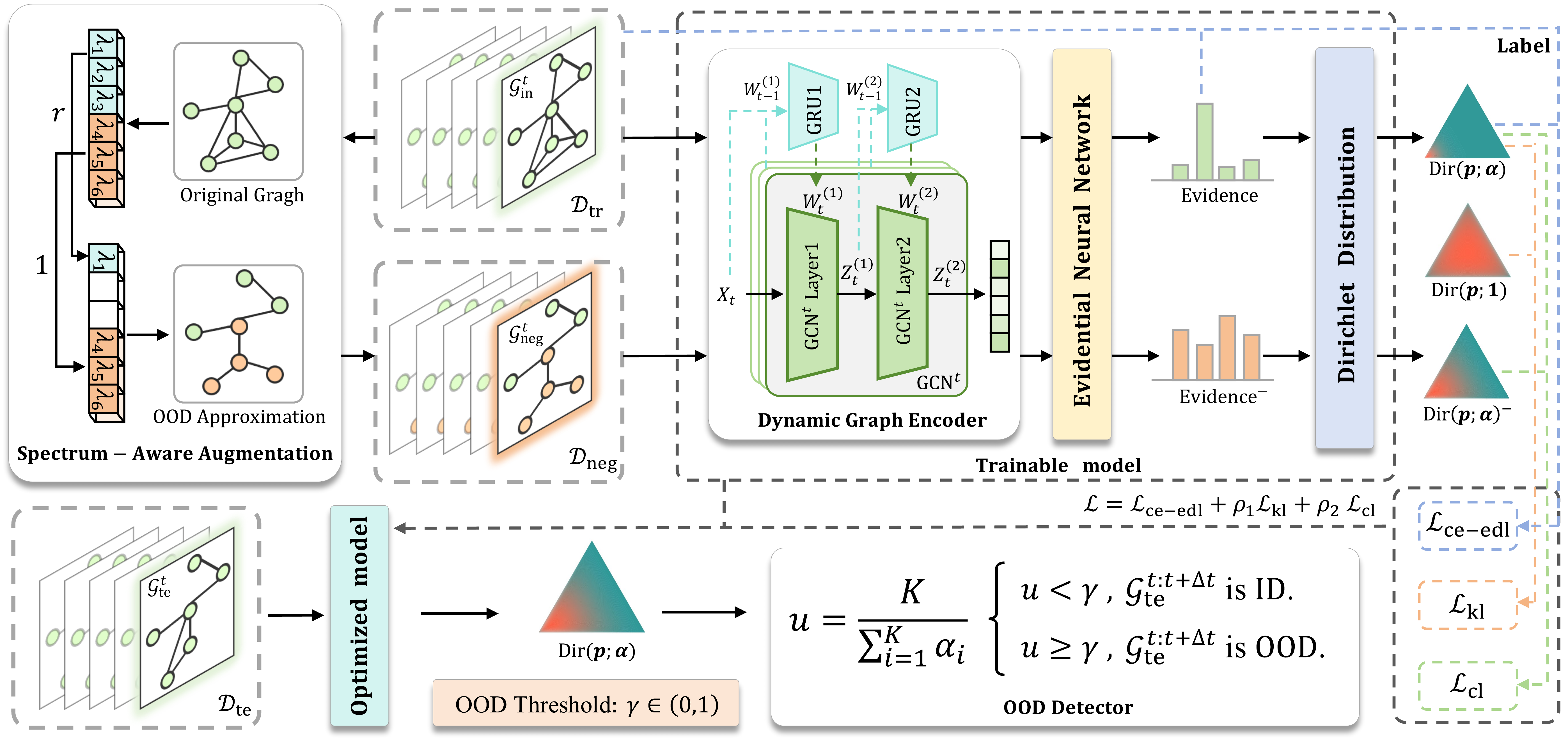} 
    \caption{An overall illustration of the proposed method, EviSEC, which follows a two-stage paradigm: (1) During training, original dynamic graph samples and OOD approximations generated via spectrum-aware augmentation are assigned different Dirichlet distributions by the dynamic graph encoder and the EDL module---sharper for ID samples and flatter for OOD approximations. The loss function enlarges the gap in their uncertainty scores. (2) During inference, the trained model accurately identifies OOD samples using a learned uncertainty threshold.}
    \label{fig:pipeline}
\end{figure}

In this section, we present a novel EDL framework for OOD detection in dynamic graphs. Our proposed model, i.e., \textbf{EviSEC} combines a flexible dynamic graph encoder (Sec.~\ref{sec:encoder}) with an evidential neural network (Sec.~\ref{sec:ecdg}) to address the limitations of single-point estimation without sacrificing ID performance. Furthermore, to mitigate score homogenization, we introduce a new spectrum-aware contrastive learning strategy (Sec.~\ref{sec:SACL}). Finally, we design an OOD detector (Sec.~\ref{sec:UQ}) based on the uncertainty scores. The overall architecture is illustrated in Fig.~\ref{fig:pipeline}.

\subsection{Dynamic Graph Encoder}\label{sec:encoder}
EviSEC adopts a dynamic graph encoder~\cite{pareja2020evolvegcn} to efficiently capture the topological structure and temporal dependencies in dynamic graphs through the evolution of GCN parameters. Specifically, we employ a GCN with weight matrix set $\mathcal{W}_t = \{\mathbf{W}^{(1)}_t, \mathbf{W}^{(2)}_t, \dots, \mathbf{W}^{(L)}_t\}$, where $\mathbf{W}^{(l)}_t$ denotes the weight matrix at $l$-th layer, to learn the node embedding matrix $\mathbf{Z}^{(l)}_t$ at each timestep $t$:
\begin{equation}\label{eq:Z}
\mathbf{Z}^{(l)}_t = 
\begin{cases}
\mathbf{X}^t, & \text{if } l = 0, \\
\sigma \left( \tilde{\mathbf{D}}^{t^{-\frac{1}{2}}}\tilde{\mathbf{A}}^t \tilde{\mathbf{D}}^{t^{-\frac{1}{2}}} \mathbf{Z}^{(l-1)}_t \mathbf{W}^{(l)}_t \right), & \text{if } l > 0,
\end{cases}
\end{equation}
where $\sigma$ is a non-linear activation function,  $\tilde{\mathbf{A}}^t$  is the adjacency matrix augmented with self-loops, and $\tilde{\mathbf{D}}^t$ denotes the diagonal degree matrix of $\tilde{\mathbf{A}}^t$. 

To naturally store historical dependencies,  $\mathbf{W}^{(l)}_t$ is regarded as a hidden state of the dynamical system. We utilize gated recurrent units (GRUs) to incorporate the layer input $\mathbf{Z}_t^{(l-1)}$ into the update of $\mathbf{W}^{(l)}_t$ over time:

\begin{equation}
\underbrace{\mathbf{W}_{t}^{(l)}}_{\text{GCN weights}} = \text{GRU}( 
\underbrace{\mathbf{Z}_t^{(l-1)}}_{\text{embeddings}}, 
\underbrace{\mathbf{W}_{t-1}^{(l)}}_{\text{GCN weights}} 
).
\end{equation}
By iteratively updating the weight matrix set from $\mathcal{W}_{t}$ to $\mathcal{W}_{t+ \Delta t}$, our dynamic graph encoder can aggregate the evolutionary patterns of the input $\mathcal{G}^{t:t + \Delta t}$ to generate the final output $f_\theta(\mathcal{G}^{t:t + \Delta t},\mathcal{W}_{t+ \Delta t}) = \mathbf{Z}_{t + \Delta t}^{(L)}$. 
To perform a specific classification task, the output of our encoder is passed through a classifier $g_\phi(\cdot)$ to generate the final prediction $\hat{\boldsymbol{y}} = \text{softmax} \left(g_\phi\left( f_\theta\left( \mathcal{G}^{t:t + \Delta t}, \mathcal{W}_{t+ \Delta t} \right) \right)\right)$.

\subsection{Evidential Neural Network}\label{sec:ecdg}
EviSEC proposes an evidential neural network to reshape the prediction $\boldsymbol{p} = [p_1, p_2, \dots, p_K]$ into a Dirichlet distribution. Thus, we can leverage the uncertainty of distribution to describe the randomness of the input, addressing the high bias and high variance issues in single-point estimation. Specifically, two loss functions are developed to guide the network to assign a sharper posterior distribution with lower uncertainty to the ID data, allowing for a robust OOD detection process. Notably, our method maintains the highest probability assigned to the target class, which preserves ID performance.

In a $K$-class classification task, since there is no prior information, we can simply assume the uniform Dirichlet distribution $\text{Dir}^{pr}(\boldsymbol{p}; \mathbf{1})$ as the prior distribution $\text{Dir}^{pr}(\boldsymbol{p}; w\boldsymbol{\beta})$, where the base rate $\boldsymbol{\beta} = \left[\frac{1}{K}, \frac{1}{K}, \dots, \frac{1}{K}\right]$ and the prior weight $ w=K $. Based on the preliminary mentioned in Sec.~\ref{sec:edl}, the posterior concentration $\boldsymbol{\alpha} = [\alpha_1, \alpha_2, \cdots, \alpha_K]$ and the class probability $\boldsymbol{p} = [p_1, p_2, \dots, p_K]$ are derived as follows,

\begin{equation}\label{ea}
\alpha_i = e_i + \beta_i w \overset{\text{Dir}^{pr}(\boldsymbol{p}; \mathbf{1})}{=}e_i + 1, \text{where} \ i = 1,2,\dots,K.
\end{equation}
\begin{equation}
p_i = \mathbb{E}_{\boldsymbol{p} \sim \text{Dir}^{po}(\boldsymbol{p};\boldsymbol{\alpha})}[p_i] =  \int p_i\text{Dir}^{po}(\boldsymbol{p}; \boldsymbol{\alpha}) \, d\boldsymbol{p}=\frac{\alpha_i}{\alpha_\text{sum}}, \text{where} \ i = 1,2,\dots,K.
\end{equation}
To update the posterior concentration  $\boldsymbol{\alpha}$, our evidential neural network includes an evidence collector $\boldsymbol{e}_{\theta,\phi}(\cdot)$ to construct the corresponding evidence vector $\boldsymbol{e} = [e_1, e_2, \dots, e_K]$:
\begin{equation}
\boldsymbol{e}_{\theta,\phi}\left( \mathcal{G}^{t:t + \Delta t} \right) = \text{exp} \left(g_\phi\left( f_\theta\left( \mathcal{G}^{t:t + \Delta t},\mathcal{W}_{t+ \Delta t} \right) \right)\right)-\textbf{1}.
\end{equation}
By subtracting all-ones vector $\textbf{1}$ from exponentiated logits, our method ensures negative logits to yield negative evidences.
To ensure that \( \boldsymbol{e}_{\theta,\phi}(\cdot) \)  produces sharp distributions with low uncertainty scores for ID samples, our network comprises two key losses: the evidential learning cross-entropy loss $\mathcal{L}_{\text{ce-edl}}$ and the Kullback-Leibler divergence loss $\mathcal{L}_{\text{kl}}$ following traditional EDL framework~\cite{sensoy2018evidential}.  $\mathcal{L}_{\text{ce-edl}}$  can be calculated as the negative log-likelihood of the predicted class probability $\boldsymbol{p}$ with respect to the one-hot label $\boldsymbol{y}= [y_1, y_2, \dots, y_K]$: 
\begin{equation}
\begin{aligned}
\mathcal{L}_{\text{ce-edl}} 
&= \frac{1}{|\mathcal{D}^{\text{tr}}|} \sum_{(\mathcal{G}^{t:t + \Delta t}, \boldsymbol{y}) \in \mathcal{D}^{\text{tr}}} 
\mathbb{E}_{\boldsymbol{p} \sim \text{Dir}^{po}(\boldsymbol{p}; \boldsymbol{\alpha})} \left[ - \sum_{i=1}^K y_i \log(p_i) \right] \\
&= \frac{1}{|\mathcal{D}^{\text{tr}}|} \sum_{(\mathcal{G}^{t:t + \Delta t}, \boldsymbol{y}) \in \mathcal{D}^{\text{tr}}} 
\sum_{i=1}^K y_i \mathbb{E}_{p_i \sim \text{Dir}^{po}(\boldsymbol{\alpha})} \left[ - \log(p_i) \right] \\
&= \frac{1}{|\mathcal{D}^{\text{tr}}|} \sum_{(\mathcal{G}^{t:t + \Delta t}, \boldsymbol{y}) \in \mathcal{D}^{\text{tr}}} 
\sum_{i=1}^K y_i \left( \psi(\alpha_{\text{sum}}) - \psi(\alpha_i) \right).
\end{aligned}
\label{ce}
\end{equation}
The digamma function $\psi(\cdot) $ is the logarithmic derivative of the Gamma function. We can see that Eq.~\ref{ce} ensures the preservation of the original ID testing performance and guides the model to assign the highest concentration to the target class. However, it may still allow $\boldsymbol{e}_{\theta,\phi}(\cdot)$ to allocate relatively high evidence to non-target classes, thereby increasing the uncertainty $u$ of ID data. To address this, we introduce $\mathcal{L}_{\text{kl}}$ to further penalize the model if it provides excessive evidence for non-target categories:

\begin{equation}
\mathcal{L}_{\text{kl}} = \frac{1}{| \mathcal{D}^{\text{tr}} |} \sum_{(\mathcal{G}^{t:t + \Delta t}, \boldsymbol{y})\in \mathcal{D}^{\text{tr}}} \text{KL} \left( \text{Dir}^{po}(\boldsymbol{p}; \boldsymbol{\hat{\alpha}}), \text{Dir}^{po}(\boldsymbol{p}; \boldsymbol{1}) \right).
\label{kl}
\end{equation}
Here $\boldsymbol{\hat{\alpha}} = \boldsymbol{y}+ (\boldsymbol{1} - \boldsymbol{y}) \odot \boldsymbol{\alpha}$, and $\odot$ represents the Hadamard product. $\boldsymbol{\hat{\alpha}}$ signifies the
adjusted concentration parameter, which reserves correct predictions while damping the contribution from irrelevant classes. 


\subsection{Spectrum-Aware Contrastive Learning}\label{sec:SACL}
In the above computational process, since only ID data is available, the model tends to assign generally lower uncertainty scores. The absence of OOD data hinders the model’s ability to learn and predict higher uncertainty scores, leading to a severe score homogenization issue and further compromising its effectiveness in distinguishing OOD data.
To mitigate this, we propose a graph spectrum-aware augmentation technique to generate negative samples as OOD approximations for dynamic graphs. Through contrastive loss, we further widen the gap in uncertainty scores between ID and OOD data.

\paragraph{\textbf{Spectrum-Aware Augmentation.}} Previous work~\cite{liu2022revisiting} demonstrates that low-frequency components in the graph spectrum capture global features (e.g., graph connectivity), while high-frequency components often reflect noise. Therefore, our technique changes the low-frequency components in the graph spectrum to generate negative samples.

Let the symmetric normalized Laplacian matrix of the adjacency matrix with self-loops $\tilde{\mathbf{A}}$  represented as $\mathbf{L} = \mathbf{I} - \tilde{\mathbf{D}}^{-1/2} \tilde{\mathbf{A}} \tilde{\mathbf{D}}^{-1/2}$, which can be eigen-decomposed as follows,
\begin{equation}
\mathbf{L} = \mathbf{U} \mathbf{\Lambda} \mathbf{U}^\top = \sum_{i=1}^N\lambda_i \boldsymbol{u}_i \boldsymbol{u}_i^\top,
\end{equation}
where $\mathbf{\Lambda} = \text{diag}(\lambda_1,\lambda_2,\dots,\lambda_N)$ is the eigenvalue diagonal matrix and $\mathbf{U} = [\boldsymbol{u}_1^\top, \boldsymbol{u}_2^\top, \dots, \boldsymbol{u}_N^\top]$ is the corresponding orthogonal eigenvector matrix. 
For simplicity's sake, we assume $0 \leq \lambda_1 \leq \lambda_2 \leq \cdots \leq \lambda_N<2 $. 
Then the graph spectrum can be partitioned into two parts: low-frequency components with eigenvalues $\boldsymbol{\lambda}_{low}=\{\lambda_1, \lambda_2, \dots, \lambda_{\lfloor N/2 \rfloor}\}$ and high-frequency components with eigenvalues $\boldsymbol{\lambda}_{high}=\{\lambda_{\lfloor N/2 \rfloor+1}, \lambda_{\lfloor N/2 \rfloor+2}, \dots, \lambda_N\}$. 

Our method generates OOD approximations by perturbing the low-frequency information with a preservation ratio r, where $0 \leq r < 1$. For the low-frequency eigenspaces $\left\{ \boldsymbol{u}_i \boldsymbol{u}_i^\top \right\}_{i=1}^{\lfloor N/2 \rfloor}$ of $\boldsymbol{\lambda}_{low}$, we only preserve the first $r$ portion and discard the remaining part. For the high-frequency eigenspaces $\left\{ \boldsymbol{u}_j \boldsymbol{u}_j^\top \right\}_{j=\lfloor N/2 \rfloor +1}^{N}$ of $\boldsymbol{\lambda}_{high}$, we keep them intact.
Based on this, the negative sample $\mathbf{L}^-_r$ is generated as follows,
\begin{equation}
\mathbf{L}^-_r = \sum_{i=1}^{\lfloor rN/2 \rfloor}\boldsymbol{u}_i \boldsymbol{u}_i^\top + \sum_{j=\lfloor N/2 \rfloor+1}^{N}\boldsymbol{u}_j \boldsymbol{u}_j^\top.
\end{equation}
\paragraph{\textbf{Contrastive Loss.}} With the above negative samples, we seek to enable the model to learn and assign higher uncertainty scores for OOD approximations, which correspond to smoother posterior distributions. To increase the gap in uncertainty scores between the sample pairs, our contrastive loss is formed as the log-likelihood between the class probability $\boldsymbol{p}^-= [p^-_1,p^-_2,\dots,p^-_K]$ of the negative sample and
the class probability $\boldsymbol{p}$ of the original input:
\begin{equation}
\mathcal{L}_{\text{cl}} 
= \frac{1}{|\mathcal{D}^{\text{tr}}|} \sum_{(\mathcal{G}^{t:t + \Delta t}, \boldsymbol{y}) \in \mathcal{D}^{\text{tr}}} \mathbb{E}_{\boldsymbol{p^-}} \left[ \sum_{i=1}^K p_i \log(p^-_i) \right].
\label{cl}
\end{equation}
With two balancing factors $\rho_1$ and $\rho_2$, the overall loss function of EviSEC is formulated as:
\begin{equation}\label{eq:loss}
\begin{aligned}
\mathcal{L} =\mathcal{L}_{\text{ce-edl}} + \rho_1  \mathcal{L}_{\text{kl}} + \rho_2 \mathcal{L}_{\text{cl}}. 
\end{aligned}
\end{equation}
\subsection{OOD Detector}\label{sec:UQ}
After our model has been optimized by Eq.~\ref{eq:loss}, EviSEC includes an OOD detector to compute the uncertainty of posterior distribution $\text{Dir}^{po}(\boldsymbol{p}; \boldsymbol{\alpha})$ as the OOD score. In specific, the uncertainty $u$ is derived from the Dirichlet concentration $\boldsymbol{\alpha}$ through Eq.~\ref{F}:
\begin{equation}
u = \frac{K}{\alpha_{\text{sum}}}, \text{where}\ \alpha_\text{sum} = \sum_{i=1}^K \alpha_i. 
\end{equation}
On the one hand, a lower $u$ corresponds to a sharper and more confident output distribution, suggesting the sample is from the ID class. On the other hand, a higher $u$ corresponds to a smoother output distribution, indicating that the sample is likely an OOD instance. Then the OOD detector can select a threshold $\gamma$ to determine whether the input is OOD according to the learned $u$.

\section{EXPERIMENT}
This section empirically validates the effectiveness of EviSEC by addressing four key research questions: \textbf{RQ1}, how does EviSEC perform compared with competitive methods? \textbf{RQ2}, how do the evidential neural network and the spectrum-aware contrastive learning enhance OOD detection in dynamic graphs? \textbf{RQ3}, how effective is our method in staying compatible with ID tasks? \textbf{RQ4}, how do key parameters affect EviSEC's performance?
\begin{table}[htbp]
\caption{Dataset details. "Time Splits" shows the data division (train/val/test). }
\centering
\begin{tabular}{l@{\hspace{12pt}}c@{\hspace{16pt}}c@{\hspace{16pt}}c@{\hspace{16pt}}c@{\hspace{16pt}}c}
\toprule
 & \textbf{\#Nodes} & \textbf{\#Edges}  & \textbf{\#Time Splits}  & \textbf{Task} \\
\midrule
BC-OTC~\cite{kumar2018rev2}     & 5,881   & 35,588    & 95 / 14 / 28   & Edge Classification \\
BC-Alpha~\cite{kumar2018rev2}   & 3,777   & 24,173    & 95 / 13 / 28   & Edge Classification \\
UCI~\cite{nr}                   & 1,899   & 59,835    & 62 / 9 / 17    & Link Prediction \\
AS~\cite{leskovec2005graphs}    & 6,474   & 13,895    & 70 / 10 / 20   & Link Prediction \\
Elliptic~\cite{weber2019anti}   & 203,769 & 234,355   & 31 / 5 / 13    & Node Classification \\
Brain~\cite{xu2019adaptive}     & 5,000   & 1,955,488 & 10 / 1 / 1     & Node Classification \\
\bottomrule
\end{tabular}
\label{tab:dataset}
\end{table}
\subsection{Experimental Setup}
\paragraph{\textbf{Datasets.}}
As summarized in Tab.~\ref{tab:dataset}, we adopt the datasets and preprocessing steps used in previous dynamic graph representation works~\cite{pareja2020evolvegcn,DEFT2023,LEDG2022}.
These include: i) \textbf{BC-OTC}, a trust-based bitcoin transaction network; ii) \textbf{BC-Alpha}, a similar bitcoin transaction network from another platform; iii) \textbf{UCI}, a student community network; iv) \textbf{AS}, a router traffic flow network; v) \textbf{Elliptic}, bitcoin transactions from the Elliptic network; and vi) \textbf{Brain}, a region connectivity network in the brain.

\paragraph{\textbf{OOD Data.}}
We extend the OOD data generation framework from recent work~\cite{wu2023energy} on static graphs  to generate OOD dynamic graphs through two strategies: i) Structure Manipulation (\textbf{SM}), which utilizes a stochastic block model to generate graphs as OOD samples, and ii) Feature Interpolation (\textbf{FI}), which applies random interpolation to create node features for OOD data.

\paragraph{\textbf{Baselines.}}
Beyond the three OOD detection families outlined in our Related Works, we also utilize several dynamic graph anomaly detection methods as baselines. These include: i) post-processing methods (MSP~\cite{hendrycks2016baseline}, ODIN~\cite{ODIN2017}, Mahalanobis~\cite{lee2018simple}), ii) energy-based methods (Energy~\cite{liu2020energy}, GNNSAFE~\cite{wu2023energy}), iii) uncertainty-based methods (Entropy~\cite{shannon1949communication}, EDL~\cite{deng2023uncertainty}, DAEDL~\cite{yoon2024uncertainty}) and iv) anomaly detection methods(NetWalk~\cite{yu2018netwalk}, TADDY~\cite{liu2021anomaly}, SLADE~\cite{lee2024slade}). 

\paragraph{\textbf{Metrics.}}
We adopt the commonly used evaluation metrics~\cite{guo2023data} for OOD detection, including: i) \textbf{AUROC}, the area under the receiver operating characteristic curve; ii) \textbf{AUPR}, area under the precision-recall curve; and iii) \textbf{FPR95}, false positive rate at 95\% true positive rate. 
Additionally, \textbf{F1}, the harmonic mean of precision and recall, is employed to evaluate ID performance.
\paragraph{\textbf{Implementation.}}
We employ grid search to determine the optimal values of three key hyper-parameters in our model: the balancing factors $ \rho_1 $ and $ \rho_2 $, and the preservation rate $ r $ in augmentation. 
The remaining parameters, including the training epochs, the number of layers $L$, hidden dimension and temporal window size $\Delta t$, are adopted from prior studies~\cite{pareja2020evolvegcn,DEFT2023}. To ensure a fair comparison, both the baselines and EviSEC are configured with identical settings. All experiments were conducted on Python 3.6.13 and NVIDIA A800 80GB GPU.

\subsection{OOD Detection Perfrmance(RQ1)}
In Tab.~\ref{tab:SM} and Tab.~\ref{tab:FI}, we present the key results of EviSEC in comparison with eleven competitive models from four families for the SM and FI OOD types, respectively. These experiments are conducted across six real-world scenarios, measured by AUROC, AUPR, and FPR95.
For the OOD type of SM, EviSEC consistently surpasses all baselines across all six datasets, boosting the average AUROC by 24.57\%, increasing the average AUPR by 25.64\%, and reducing the average FPR95 by 34.65\%. For the OOD type of FI, the corresponding changes are 24.08\%, 25.22\%, and 37.39\%, respectively. These results demonstrate that our model exhibits strong adaptability to different OOD types. 
In both OOD types, we are pleased to observe exceptionally satisfactory performance on the BTC-OTC and BC-Alpha datasets. Specifically, we achieved nearly perfect AUROC and AUPR scores, as well as extremely low FPR95, significantly surpassing the performance of other models. These results reinforce the superiority of EviSEC for OOD detection in dynamic graphs.

\begin{table*}[t]
\caption{OOD detection results for the OOD type of structure manipulation (\textbf{SM)} evaluated  by AUROC/AUPR/FPR95 on six datasets. The best performance is \textbf{bolded}, with the runner-up \underline{underlined}.}
\resizebox{\textwidth}{!}{
\begin{tabular}{@{}c||ccc|ccc|ccc|ccc|ccc|ccc@{}}
\toprule
\multirow{2}{*}{\begin{tabular}{c} \diagbox{Baseline}{Dataset} \end{tabular}} 
                  & \multicolumn{3}{c|}{BC-OTC} & \multicolumn{3}{c|}{BC-Alpha} & \multicolumn{3}{c|}{UCI} & \multicolumn{3}{c|}{AS} & \multicolumn{3}{c|}{Elliptic} & \multicolumn{3}{c}{Brain} \\\cmidrule(lr){2-19} 
                  & \multicolumn{18}{c}{AUROC ↑ \hspace{10pt}   AUPR ↑  \hspace{10pt}  FPR95 ↓} \\
\midrule   
NetWalk    & 52.38 & 48.06 & 92.80 & 48.62 & 52.70 & 98.40 & 48.67 & 46.02 & 98.68 & 52.71 & 56.31 & 94.20 & 38.97 & 46.58 & 91.30 & 54.38 & 52.60 & 100.0 \\
TADDY      & 77.20 & 65.38 & 98.07 & 64.28 & 59.05 & 95.20 & 54.80 & 58.69 & 95.18 & 56.89 & 55.73 & 88.67 & 52.61 & 54.98 & 93.20 & 56.80 & 59.08 & 100.0 \\
SLADE      & 74.35 & 64.30 & 94.26 & 66.40 & 60.50 & 92.80 & 59.80 & 50.20 & 94.19 & 51.20 & 56.98 & 89.20 & 48.80 & 54.70 & 95.20 & 51.30 & 55.47 & \underline{95.58} \\
\midrule
MSP        & 48.65 & 52.82 & 92.55 & 45.79 & 48.63 & 91.50 & 54.38 & 46.01 & 97.20 & 52.76 & 47.39 & 94.20 & 48.64 & 40.43 & 90.25 & 44.87 & 42.15 & 99.80 \\
ODIN       & 55.15 & 42.63 & 88.88 & 55.26 & 49.07 & 100.0 & 41.68 & 39.92 & 96.68 & 56.08 & 45.25 & 98.34 & 48.77 & 50.62 & 92.57 & 48.83 & 42.39 & 100.0 \\
Mahalanobis& 61.44 & 34.88 & 96.80 & 43.21 & 39.34 & 92.34 & 40.68 & 38.72 & 94.29 & 57.26 & 49.77 & 97.04 & 49.56 & 52.03 & 98.62 & 36.20 & 32.17 & 98.65 \\
\midrule
Energy     & 58.80 & 31.88 & 88.11 & 50.33 & 52.68 & 99.19 & 50.68 & 55.48 & 92.70 & 57.21 & 54.23 & 95.46 & 50.69 & 52.70 & 98.37 & 34.67 & 48.94 & 99.27 \\
GNNSAFE   & 87.02 & 85.57 & 85.32 & \underline{89.66} & \underline{86.34} & \underline{78.20} & 62.41 & \underline{66.34} & \underline{90.04} & 67.08 & 59.71 & \underline{87.62} & 64.58 & 60.22 & 96.55 & 72.12 & \underline{71.17} &  98.87\\
\midrule
Entropy    & 45.15 & 23.48 & 99.09 & 50.55 & 55.18 & 100.0 & 60.27 & 50.66 & 99.21 & 55.39 & 60.22 & 92.44 & 52.66 & 55.71 & 98.06 & 61.62 & 52.33 & 99.80 \\
EDL        & 61.83 & 28.35 & 98.40 & 55.33 & 54.08 & 100.0 & 55.18 & 47.02 & 96.55 & 58.77 & 62.72 & 90.58 & 54.87 & 50.68 & 97.54 & 65.99 & 66.60 & 99.40 \\
DAEDL      & \underline{90.12} & \underline{88.33} & \underline{69.88} & 85.34 & 80.20 & 90.68 & \underline{70.22} & 64.09 & 90.27 & \underline{74.48} & \underline{69.85} & 88.36 & \underline{66.53} & \underline{68.97} & \underline{80.65} & \underline{74.39} & 70.17 & 96.31 \\
\midrule
EviSEC     & \textbf{96.39} & \textbf{94.52} &\textbf{13.47 }& \textbf{97.14} & \textbf{92.30} & \textbf{13.86} & \textbf{72.09} & \textbf{68.84} & \textbf{82.77} &\textbf{78.39} & \textbf{76.50}& \textbf{80.46} & \textbf{70.61} &\textbf{72.35} & \textbf{74.68} & \textbf{76.73} & \textbf{73.51} & \textbf{92.64} \\

\bottomrule

\end{tabular}
}
\label{tab:SM}
\end{table*}
\begin{table*}[t]
\caption{OOD detection results for the OOD type of feature interpolation (\textbf{FI}) evaluated  by AUROC/AUPR/FPR95 on six datasets. The best performance is \textbf{bolded}, with the runner-up \underline{underlined}.}
\resizebox{\textwidth}{!}{
\begin{tabular}{@{}c||ccc|ccc|ccc|ccc|ccc|ccc@{}}
\toprule

\multirow{2}{*}{\begin{tabular}{c} \diagbox{Baseline}{Dataset} \end{tabular}} 
                  & \multicolumn{3}{c|}{BC-OTC} & \multicolumn{3}{c|}{BC-Alpha} & \multicolumn{3}{c|}{UCI} & \multicolumn{3}{c|}{AS} & \multicolumn{3}{c|}{Elliptic} & \multicolumn{3}{c}{Brain} \\\cmidrule(lr){2-19} 
                  & \multicolumn{18}{c}{AUROC ↑ \hspace{10pt}   AUPR ↑  \hspace{10pt}  FPR95 ↓} \\
\midrule   
NetWalk    & 57.06 & 58.54 & 91.78 & 60.45 & 65.12 & 99.20 & 50.86 & 48.34 & 90.60 & 56.27 & 50.20 & 88.71 & 42.08 & 45.32 & 99.80 & 42.48 & 37.08 & 100.0 \\
TADDY      & 79.68 & 64.43 & 94.07 & 68.31 & 66.61 & 96.08 & 56.76 & 52.63 & 94.71 & 55.16 & 51.20 & 85.64 & 50.65 & 55.08 & 96.20 & 52.99 & 57.04 & 100.0 \\
SLADE      & 76.81 & 64.09 & 93.26 & 65.86 & 58.06 & 94.52 & 54.32 & 49.06 & 90.35 & 50.34 & 55.98 & 90.70 & 52.74 & 58.62 & 94.64 & 47.03 & 55.93 & 98.40 \\
\midrule
MSP        & 43.04 & 38.80 & 94.38 & 46.28 & 38.98 & 94.34 & 33.74 & 42.58 & 98.77 & 51.20 & 48.29 & 99.25 & 44.28 & 39.38 & 88.64 & 34.18 & 33.66 & 100.0 \\
ODIN       & 50.26 & 47.87 & 90.32 & 54.20 & 50.84 & 99.80 & 46.20 & 48.07 & 98.82 & 58.91 & 47.30 & 96.53 & 50.24 & 59.26 & 94.40 & 50.07 & 56.30 & 99.25 \\
Mahalanobis& 45.20 & 51.35 & 98.20 & 45.84 & 42.09 & 95.63 & 43.82 & 44.24 & 93.29 & 56.98 & 51.28 & 95.77 & 48.53 & 54.08 & 89.60 & 44.31 & 42.07 & 98.65 \\
\midrule
Energy     & 60.28 & 59.60 & 72.37 & 51.20 & 59.61 & 95.44 & 55.21 & 56.38 & 90.52 & 54.30 & 52.27 & 94.70 & 58.89 & 60.58 & 94.20 & 48.67 & 50.38 & 97.27 \\
GNNSAFE    & 70.05 &  \underline{86.79} & 68.30 & \underline{87.25} & \underline{88.37} & 68.46 & 64.09 & \underline{68.27} & \underline{89.42} & 66.57 & 64.35 & \textbf{80.29} & 66.80 & 58.92 & 94.38 & 63.44 & 60.18 & \underline{90.43} \\
\midrule
Entropy    & 42.95 & 26.83 & 100.0 & 47.23 & 50.35 & 100.0 & 55.69 & 54.31 & 98.70 & 52.17 & 48.69 & 88.64 & 59.07 & 62.18 & 96.50 & 70.02 & 71.17 & 96.56 \\
EDL        & 53.81 & 38.03 & 95.42 & 59.30 & 52.18 & 100.0 & 48.67 & 49.72 & 94.21 & 64.38 & 64.80 & \underline{82.50} & 61.20 & 64.52 & 90.20 & 64.20 & 68.17 & 93.57 \\
DAEDL      & \underline{80.54} & 86.28 & \underline{54.31} & 80.14 & 74.58 & \underline{65.31} & \underline{65.28} & 64.97 & 91.25 & \underline{70.07} & \underline{69.57} & 86.25 & \underline{68.34} & \underline{66.30} & \underline{78.57} & \underline{72.18} & \underline{73.15} & 96.51 \\
\midrule
EviSEC     & \textbf{97.29} & \textbf{98.51} &\textbf{5.22}& \textbf{96.40} & \textbf{94.44} & \textbf{10.27} & \textbf{70.91} & \textbf{72.08} & \textbf{80.16} &\textbf{72.58} & \textbf{74.25}& 84.20 & \textbf{72.15} &\textbf{70.25} & \textbf{58.68} & \textbf{74.21} & \textbf{76.47} & \textbf{89.40} \\

\bottomrule
\end{tabular}
}
\label{tab:FI}
\end{table*}

\subsection{Ablation Study(RQ2)}
To systematically evaluate the contribution of core components ($\mathcal{L}_{\text{ce-edl}}$, $\mathcal{L}_{\text{kl}}$ and $\mathcal{L}_{\text{cl}}$) in our framework, we conducted comprehensive ablation studies across six benchmark datasets. Through progressive removal of individual loss components from the complete EviSEC architecture, we quantitatively assessed their respective impacts on OOD detection capability measured by \textbf{AUROC}. As presented in Tab.~\ref{tab:ablation_study}, the full configuration demonstrates significant superiority over all ablated variants, with particular note to the Elliptic dataset where we observe a substantial absolute performance improvement of 17.41\%. This empirical validation confirms that the synergistic combination significantly enhances OOD detection performance.
\begin{table}[ht]
\centering
\caption{Ablation study results for the SM OOD type across six datasets. \checkmark\kern-1.6ex\raisebox{0.5ex}{\rotatebox[origin=c]{125}{\textbf{-}}} means the variant with random negative edge sampling augmentation.}
\begin{tabular}{c@{\hspace{12pt}}c@{\hspace{12pt}}c@{\hspace{12pt}}c@{\hspace{12pt}}c@{\hspace{12pt}}c@{\hspace{12pt}}c@{\hspace{12pt}}c@{\hspace{12pt}}c@{\hspace{12pt}}c} 
\toprule
$\mathcal{L}_{\text{ce-edl}}$ & $\mathcal{L}_{\text{kl}}$ & $\mathcal{L}_{\text{cl}}$ & BC-OTC & BC-Alpha  & UCI & AS & Elliptic & Brain \\
\midrule
\midrule
- & \checkmark & \checkmark          & 86.24 & 85.14 & 66.28 & 65.80 & 58.32 & 68.45 \\
\checkmark & - & \checkmark          & \underline{95.58} & \underline{95.28} & \underline{70.60} & 72.87 &\underline{68.24} & \underline{73.12} \\
\checkmark & \checkmark & -          & 90.17 & 93.69 & 68.07 & \underline{74.10} & 65.78 & 69.56 \\
\checkmark & \checkmark & \checkmark\kern-1.6ex\raisebox{0.5ex}{\rotatebox[origin=c]{125}{\textbf{-}}}          & 91.08 & 92.26 & 69.31 & 73.54 & 67.92 & 70.27 \\
\midrule
\checkmark & \checkmark & \checkmark & \textbf{96.39} & \textbf{97.14} & \textbf{72.09} & \textbf{78.39} & \textbf{70.61} & \textbf{76.73} \\

\bottomrule
\end{tabular}

\label{tab:ablation_study}
\end{table}

\subsection{ID Performance(RQ3)}
\begin{figure}[h]
    \centering
    \includegraphics[width=0.8\textwidth]{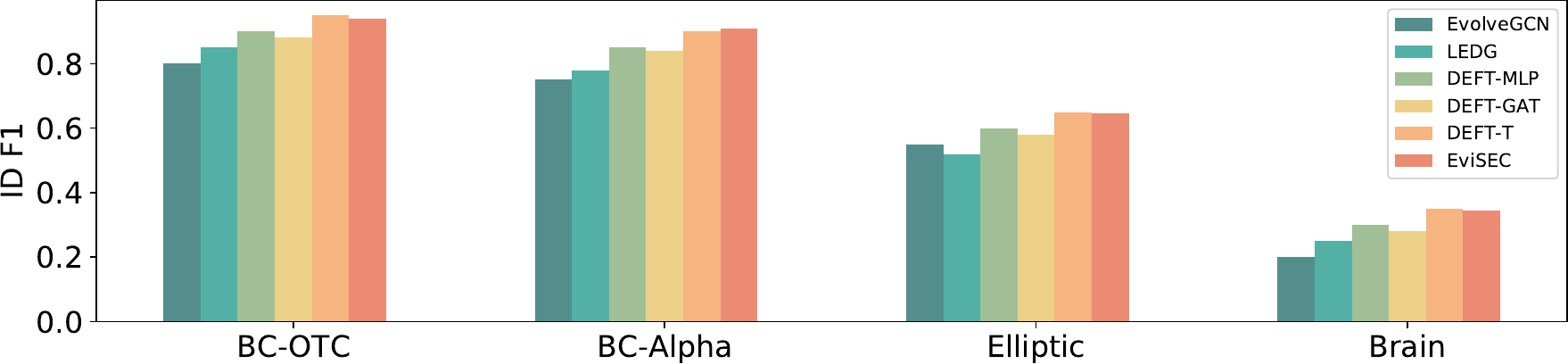} 
    \caption{In-Distribution performance of classification tasks with \textbf{F1} scores on the y-axis.}
    \label{fig:id}
\end{figure}

As mentioned in Sec.~\ref{sec:ecdg}, EviSEC does not alter the value of the maximum component in the predictive distribution. Therefore, when comparing ID performance measured by \textbf{F1} scores with representation learning methods such as EvolveGCN~\cite{pareja2020evolvegcn}, LEDG~\cite{LEDG2022}, DEFT-MLP, DEFT-GAT and DEFT-T~\cite{DEFT2023}, our approach still achieves performance close to (or even surpassing) state-of-the-art levels (see Fig.~\ref{fig:id}). This demonstrates that our method does not sacrifice ID task performance while enhancing OOD detection capabilities.

\subsection{Parameter Sensitivity(RQ4)}
\begin{figure*}[ht]
 \centering

 \subfigure{
  \includegraphics[width=6cm, height=4.5cm]{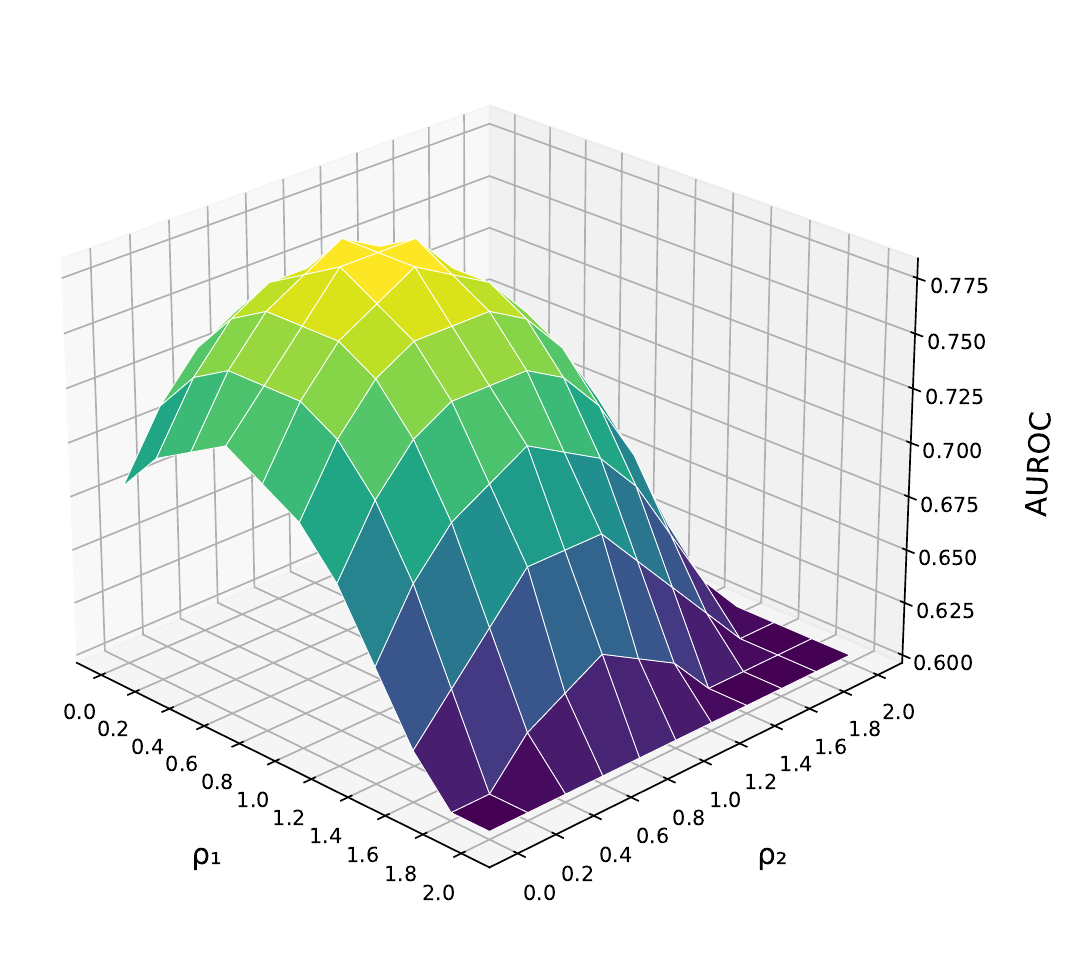}
 }
 \hspace{0.5cm} 
 \subfigure{
  \includegraphics[width=4.8cm, height=3.6cm]{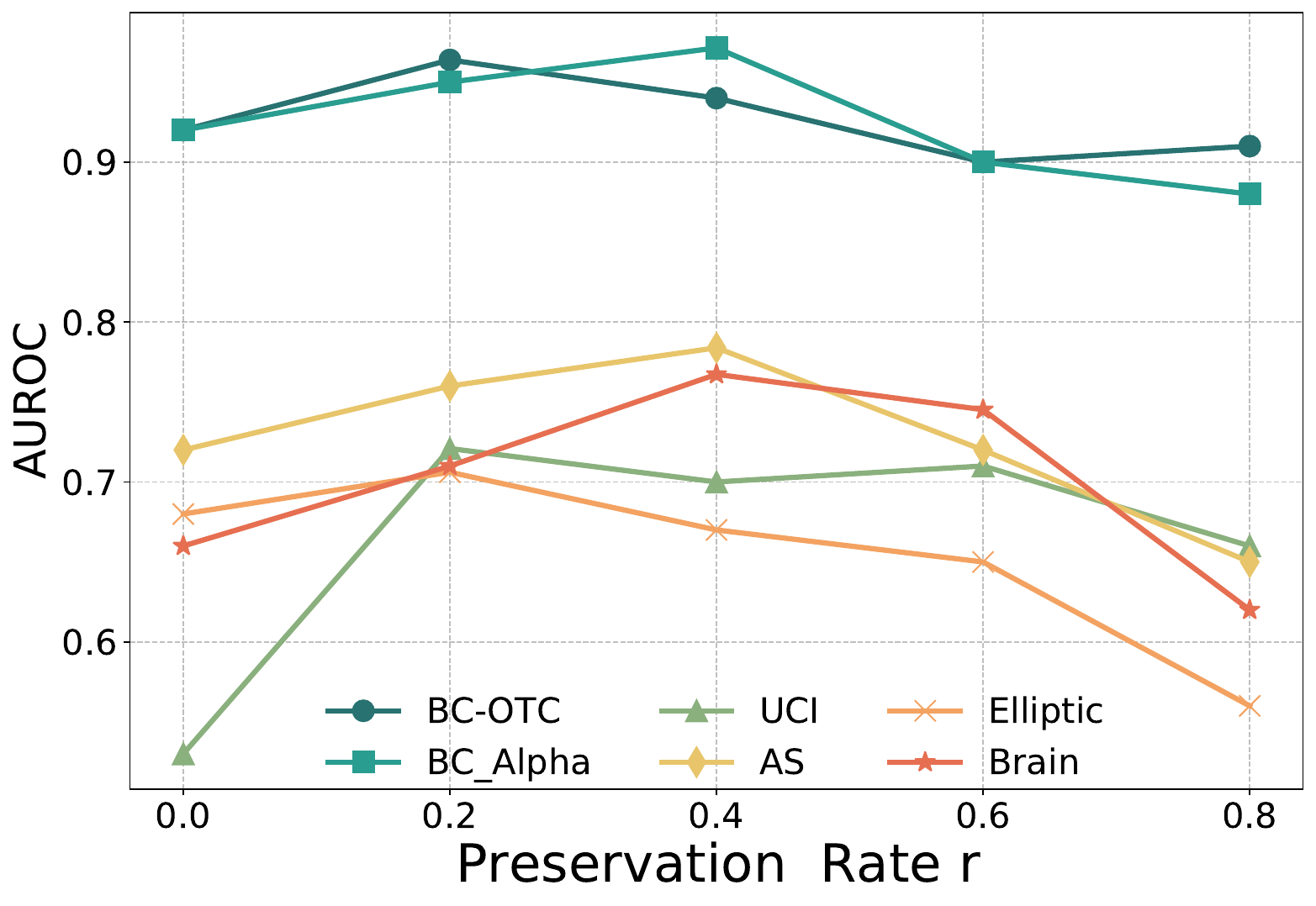}
 }
\caption{Hyper-parameter sensitivity of EviSEC.}

\label{fig:sensitivity}
\end{figure*}

Fig.~\ref{fig:sensitivity} shows the hyper-parameter sensitivity of EviSEC measured by AUROC in terms of the balancing factors $\rho_1$ and $\rho_2$, and the preservation rate $r$ in augmentation.  Specifically, $ \rho_1, \rho_2 $ range from 0.2 to 2.0 with a step size of 0.2 on the AS dataset, and $ r \in \{0, 0.2, 0.4, 0.6, 0.8\} $ on six datasets. We observe that setting them to moderate values (e.g., $\rho_1 \in \{0.4, 0.6, 0.8\}$, $\rho_2 \in \{0.6, 0.8, 1.0\}$, $ r \in [0.2,0.4] $) usually achieves optimal performance. We can conclude that: i) $\mathcal{L}_{\text{ce-edl}}$ and $\mathcal{L}_{\text{cl}}$ have a greater impact than $\mathcal{L}_{\text{kl}}$; and ii) a smaller r, i.e., retaining less low-frequency information, tends to generate more effective OOD approximations.

\section{Conclusion}
In this work, we propose EviSEC, an innovative and effective OOD detection framework comprising two key modules. Specifically, we propose an evidential neural network that uses the uncertainty of posterior distribution to explain the randomness of the input, thereby mitigating the high bias and high variance issues associated with single-point estimation. Moreover, we design a spectral-aware contrastive learning module to generate OOD approximations for enlarging the score gap between the ID and OOD data, effectively mitigating the issue of score homogenization.
Empirical evaluation demonstrates that our model outperforms others across various datasets for OOD detection in dynamic graphs.

\begin{credits}
\subsubsection{\ackname} 

This work is supported by the National Key Research and Development Program of China (NO.2022YFB3102200) and the National Natural Science Foundation of China (No.62402491). 

\end{credits}
%
%
%


\end{document}